\documentclass{article}
\usepackage{spconf,amsmath,graphicx}
\usepackage{hyperref}

\newcommand{\norm}[1]{\left\lVert#1\right\rVert}
\def\x{{\mathbf x}}
\def\w{{\mathbf w}}
\def\L{{\cal L}}
\def\Q{{\text Q}}
\def\X{{\text X}}
\usepackage[table,xcdraw]{xcolor}

\title{BOOSTING RARE BENTHIC MACROINVERTEBRATES TAXA IDENTIFICATION WITH ONE-CLASS CLASSIFICATION}
\name{Fahad Sohrab,\thanks{$\copyright$ 2020 IEEE. Personal use of this material is permitted. Permission from IEEE must be obtained for all other uses, in any current or future media, including reprinting/republishing this material for advertising or promotional purposes, creating new collective works, for resale or redistribution to servers or lists, or reuse of any copyrighted component of this work in other works.} Jenni Raitoharju$^{\dag}$\thanks{Thanks to Academy for Finland project 324475 for funding the work of Jenni Raitoharju. Emails: \{fahad.sohrab, jenni.raitoharju\}@tuni.fi.}}
\address{Faculty of Information Technology and Communication Sciences, Tampere University, Tampere, Finland\\
$^{\dag} $Programme for Environmental Information, Finnish Environment Institute, Jyv\"askyl\"a, Finland\\
}
\begin{document}
%
\maketitle
\begin{abstract}
Insect monitoring is crucial for understanding the consequences of rapid ecological changes, but taxa identification currently requires tedious manual expert work and cannot be scaled-up efficiently. Deep convolutional neural networks (CNNs), provide a viable way to significantly increase the biomonitoring volumes. However, taxa abundances are typically very imbalanced and the amounts of training images for the rarest classes are simply too low for deep CNNs. As a result, the samples from the rare classes are often completely missed, while detecting them has biological importance. In this paper, we propose combining the trained deep CNN with one-class classifiers to improve the rare species identification. One-class classification models are traditionally trained with much fewer samples and they can provide a mechanism to indicate samples potentially belonging to the rare classes for human inspection. Our experiments confirm that the proposed approach may indeed support moving towards partial automation of the taxa identification task.

\end{abstract}
\vspace{-5pt}
\section{Introduction}
\label{sec:intro}
To understand the consequences of climate change and other anthropogenic changes in different aquatic ecosystems, it is crucial to widely monitor different animal groups. 
Also international environmental legislation, such as the EU Water Framework Directive (WFD) \cite{EUdirective2000},
acknowledges the task of monitoring aquatic ecosystems. Since changes in the abundances of benthic macroinvertebrate species can provide an early warning sign of environmental problems in aquatic ecosystems, they are widely used as indicating factors in WFD-compliant ecological status assessment and environmental decision making \cite{buchner2019analysis, sun2019freshwater}. At the same time, they have been identified also as one of the most difficult groups to be monitored \cite{poikane2016benthic}. The task currently requires tedious manual expert work making it expensive, time-consuming, and error-prone. The recent advances in machine learning, especially deep convolutional neural networks (CNNs), provide a viable way to scale-up monitoring and provide faster information for environmental decision making. In the future, the samples of benthic macroinvertebrates may be imaged with an automated imaging device and then identified using a deep learning model trained with a sample dataset. 

The overall accuracy obtained by automatic identification of benthic macroinvertebrates is approaching human expert level \cite{arje2017human} and, already in the near future, it may be possible to use machines to handle the majority of the samples, while human experts manually identify only the difficult and interesting cases, such as specimens potentially belonging to rare species. A major challenge that needs to be addressed is induced by the very imbalanced taxa abundances. For some rare species, the number of training images is simply too low for a deep CNN and, as a result, the identification often fails. This problem is largely overlooked in the recent works \cite{arje2017human, raitoharju2019confidences} that consider only the overall identification accuracy. The low number of misclassified specimens from rare species hardly affects the overall accuracy, while they are important for monitoring biodiversity. In this paper, we propose a mechanism that can indicate a reasonably-sized subset of specimens as potential samples of rare species for human expert inspection. To this end, we propose combining the trained deep CNN with one-class classifiers. One-class classifiers are traditionally trained with much fewer samples than deep networks and our experimental results support the assumption that they can help in detecting samples from the rare species.

\section{Related work}\label{sec:relatedwork}
\subsection{Machine learning in biomonitoring}

Machine learning is rapidly gaining recognition as a promising tool for many biomonitoring applications, such as identifying fish species \cite{santos2019fish}, forest surveillance \cite{wyniawskyj2019forest}, or monitoring Arctic flowering seasons \cite{arje2019flower}. In this paper, we concentrate on benthic macroinvertebrate identification. Nevertheless, main challenges are similar for most biomonitoring applications and the solutions may be easily applied on other applications. For example, the identification task is very fine-grained. For a non-expert it may be hard to see any difference between similar species. At the same time, the intra-class variance may be large due to different development stages \cite{raitoharju2018benchmark}. Taxa distributions in the nature, and thus also the available reference datasets, are very imbalanced \cite{arje2017human}. Furthermore, some taxonomists continue to object the shift toward automated methods due to different doubts and fears \cite{nonscientific_factors}. The last problem may me eased by providing better mechanisms for dividing the identification task between machines and human experts in such a manner that the machine first handle only the most routine-like cases \cite{raitoharju2019confidences}.

Efforts to develop automated taxa identification techniques have developed from using handcrafted features with shallow networks \cite{lytle2010benthic, kiranyaz2011classification} towards using deep neural networks, which operate on images as inputs \cite{arje2017human, raitoharju2019confidences}. A major challenge with deep neural networks is the need of huge amounts of training data. This had lead to efforts to create imaging devices capable of providing high quality images with minimal manual effort \cite{raitoharju2018benchmark, arje2020biodiscover}. 
Nevertheless, the existing datasets, such as FIN-Benthic2 \cite{arje2017human} used in this paper, have very imbalanced classes. The smallest taxa simply do not provide enough information for training deep neural networks. However, such rare taxa and changes in their abundances may be biologically and environmentally interesting. The performance of the deep neural networks for the very rare species may be enhanced, e.g., by data-augmentation \cite{raitoharju2016data} or special loss functions \cite{huang2016loss}, but also these approaches tend to overfit to the few training samples and do not generalize well for unseen samples.  In this paper, we suggest combining one-class classifiers with the trained deep neural network to provide an additional mechanism for detecting samples potentially belonging to the rare classes for human inspection. 

\vspace{-10pt}
\subsection{One-class classification} \label{ssec:occ}
The main idea in one-class classification is to create a representative model of a class of interest, typically called target class, using data from this class only. During inference, the model is used to predict whether unseen samples belong to the target class or are outliers. We denote the target data as $\X = [\x_1, ..., \x_n]$, where $n$ is the number of target items and $\x_i$ are $D$-dimensional vectors.  One-Class Support Vector Machine (OC-SVM) \cite{scholkopfu1999sv} basically separates all the data points from the origin and maximizes the distance from this hyperplane to the origin:
\begin{eqnarray}\label{OCSVM}
\min_{\w, \xi_i, \rho}  \quad & \frac{1}{2} \norm{\w}^2 + \frac{1}{C n} \sum_{i=1}^{n} \xi_i - \rho \nonumber\\
\textrm{s.t.} \quad & \w*\x_i \ge \xi_i - \rho, \:\: \forall i\in\{1,\dots,n\}  \nonumber\\
 \quad & \xi_i \ge 0, \:\: \forall i\in\{1,\dots,n\},
\end{eqnarray}
where $\w$ is a weight vector, slack variables $\xi_i$ allow some data points to lie within the margin, and hyper-parameter $C$ sets an upper bound on the fraction of training samples allowed within the margin and a lower bound on the number of training samples used as Support Vector. 

Another classical one-class classification method is Support Vector Data Description (SVDD) \cite{tax2004support}. An SVDD model is trained by forming the smallest hypersphere which includes all the target data. SVDD minimized the following function:
\begin{eqnarray}\label{erfuncSVDD}
\min \quad & F(R,\textbf{a}) = R^2 + C \sum_{i=1}^{n} \xi_i \nonumber\\
\textrm{s.t.} \quad & \| {\mathbf{x}_i} - \mathbf{a} \|^2_{2} \le R^2 + \xi_i, \:\: \forall i\in\{1,\dots,n\}, \nonumber \\
\quad & \xi_i \ge 0, \:\: \forall i\in\{1,\dots,n\},
\end{eqnarray}  
where $R$ is the radius, $\textbf{a}$ is the center of hypersphere, $\xi_i$ are slack variables allowing some training samples to be left outside the hypersphere, and hyper-parameter $C$ controls the amount of allowed outliers. Both OC-SVM and SVDD can be solved in one step using Lagrange multipliers.

A recent extension of SVDD, Subspace Support Vector Data Description (S-SVDD) \cite{sohrab2018subspace} maps the data to an optimised $d$-dimensional subspace suitable for one-class classification as $\Q\x_i$. S-SVDD is solved iteratively alternating the steps of solving SVDD in the current subspace and improving the subspace projection $\Q$. The second step computes the gradient of Lagrangian of Eq. \eqref{erfuncSVDD}, $\Delta \L$, and updates $\Q$ as 
\begin{equation}\label{ssvddQ}
\Q = \Q - \eta(\Delta \L + \beta \Delta\Psi),
\end{equation}
where $\Psi = \text{Tr}(\Q\X\lambda \lambda^T \X^T\Q^T ) $ is an additional regularization term enforcing more variance, $\beta$ is a weight for it, and $\eta$ is a learning rate. Different values for $\lambda$ result in different versions of S-SVDD. In this paper, we use unregularized version (i.e. $\lambda_i = 0$) denoted as S-SVDD and two regularized versions S-SVDDr1 with $\lambda_i = 1$ and SVDDr2, where $\lambda$ is used to select only the support vectors. 

\section{Proposed system}

 \begin{figure*}[th]
	\centering
	\includegraphics[width=1\textwidth]{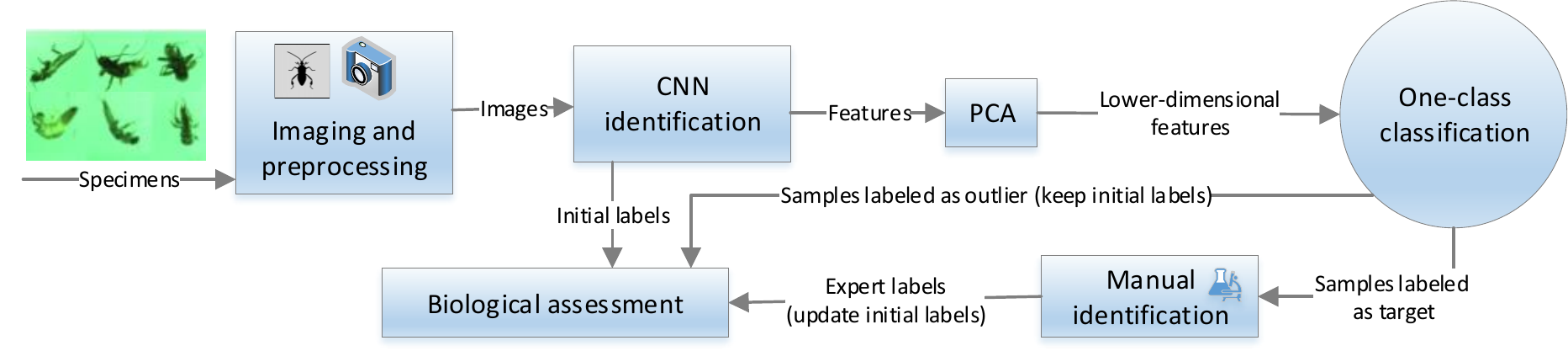}
	\caption{The proposed taxa identification pipeline}
	\label{fig:bugsystem}
\end{figure*}

Our work aims at allowing to move from fully manual taxa identification of benthic macroinvertebrates to a semi-automated approach, where a trained machine learning model can handle most of the specimens, while the human experts can concentrate on difficult and potentially most interesting cases. As our starting point, we assume the typical scenario where we have a trained deep neural network model that gives a satisfactory overall accuracy, while it fails to correctly identify specimens from rare species, which have biological/environmental importance. We propose a mechanism that can be used together with the deep neural network to pin-point specimens that potentially belong to the rare species for human expert inspection.

The proposed general framework is shown in Fig.~\ref{fig:bugsystem}. In the first phase, collected macroinvertebrate samples are imaged and the images are preprocessed as needed (e.g., normalization, resizing). The images are fed to a trained deep neural network for initial identification. Features extracted from the second last layer of the network are projected to a lower dimensionality with Principal Component Analysis (PCA) to make the one-class models smaller and more focused on the key features. The PCA-processed features are classified using a one-class classifier. Finally, the specimens which are classified to the target class are re-identified by a human expert, while otherwise the initial CNN identification is used in the subsequent biological assessment of the results. Note that the experts use the actual specimens with a microscopic analysis, while the machine learning components rely on images and features extracted from the images.

As one-class classifiers use only target class data for training, they may not be able to accurately distinguish unseen target samples from outliers, which have a high similarity with the target class. However, this may be even a benefit in our application. Trying to separate target samples from very similar outliers is naturally error-prone. Therefore, it is better to direct also these unclear cases for expert identification instead of trying to build the model as accurate as possible. In general, our goal is to detect as many samples from the target class as possible with the minimum amounts of overall samples that require manual identification. However, it is not straightforward how to evaluate the performance of different one-class classifiers on the given task. Depending on biomonitoring goals and importance of the target class, it may vary how much human effort is acceptable to maximize the number of detected target class specimens.   
\vspace{-5pt}
\section{Experimental setup and results}
\label{resultsexperiments}

\subsection{Dataset}
\label{ssec:dataset}
We used FIN-Benthic2 dataset \cite{arje2017human} in our experiments. The dataset is publicly available and consists of 460004 images of 9631 benthic macroinvertebrate specimens belonging to 39 different taxa. The number of images per taxon varies from 490 to 44240 making the dataset very imbalanced. The images are of varying size and in PNG-format. FIN-Benthic2 provides 10 different data splits for training, validation, and testing. Each split has been formed so that the images of a single specimen (max 50) are in the the same set (train/validation/test). In this paper, we consider only image-based identification and we leave it for future work to investigate how to exploit the fact that we actually have several images corresponding to the same specimen.  We used Split 1 as our data splitting. 

For one-class classification, we selected three different taxa, \textit{Capnopsis schilleri}, \textit{Nemoura cinerea}, and \textit{Leuctra nigra}, as our target classes. Each of these taxa is rare and VGG16 has poor performance on them. The target classes were selected as a proof-of-concept, not based on their environmental importance. The image numbers for the selected classes are shown in Table~\ref{Tab:amounts}. 

\vspace{-5pt}
\begin{table}[ht]
\small
\centering
\caption{Image numbers in Split 1 of FIN-Benthic2 dataset}
\label{Tab:amounts}

\begin{tabular}{l|l|l|l|}
 & Train & Validation & Test\\
\hline
\textit{Capnopsis schilleri} 
& 600 & 100    & 350\\
\textit{Nemoura cinerea} 
& 650 & 100    & 50\\
\textit{Leuctra nigra} 
&1100     & 50   & 200\\
\hline
Whole dataset & 321407 & 45912 & 92685\\

\end{tabular}
\end{table}

\vspace{-5pt}

\begin{table*}[ht]
  \centering
   \caption{One-class classifier results for different target species}
\small
\begin{tabular}{|l||l|l|l|l||l|l|l|l||l|l|l|l|}
\cline{2-13}
\multicolumn{1}{l||}{} & \multicolumn{4}{c||}{\textit{Capnopsis schilleri}}  &\multicolumn{4}{c||}{\textit{Nemoura cinerea}} &\multicolumn{4}{c|}{\textit{Leuctra nigra}}\\
\cline{2-13}
\multicolumn{1}{l||}{}       &TPR & GM & TP &TP+FP &TPR & GM & TP &TP+FP &TPR & GM & TP &TP+FP\\
\cline{2-13}
\multicolumn{13}{l}{CNN classification}  \\
\hline
VGG16  & 0.046 & 0.214 & 16   & 101     
       & 0.020 & 0.141 & 1    & 39     
			 & 0.170 & 0.412 & 34   & 174 \\
\hline
\multicolumn{13}{l}{Linear one-class classification}  \\
\hline
OC-SVM & 0.906  & 0.613 & 317 & 54367  
       & 0.660 & 0.357 & 33   & 74739  
			 & 0.625 & 0.437 & 125  & 64304 \\
SVDD   & 0.346 & 0.586 & 121  & 701                             
       & 0.280 & 0.525 & 14   & 1422  
			 & 0.730 & 0.832 & 146  & 4860  \\ 
S-SVDD             
       & 0.557 & 0.740 & 195 & 1893                            
			 & 0.480 & 0.676 & 24  & 4385
  		 & 0.805 & 0.838 & 161 & 11910 \\  
S-SVDDr1               
       & 0.609 & 0.773 & 213 & 1977                            
			 & 0.340 & 0.567 & 17  & 5209                      
			 & 0.805 & 0.837 & 161 & 12103  \\   
S-SVDDr2               
       & 0.706 & \textbf{0.825} & 247 & 3573                            
			 & 0.560 & \textbf{0.702} & 28  & 11178
			 & 0.855 & \textbf{0.876} & 171 & 9625 \\
\hline
\multicolumn{13}{l}{Non-linear one-class classification}  \\
\hline
OC-SVM 
       & 0.034 & 0.185 & 12  & 87                              
			 & 0.000 & 0.000 & 0   & 51                
			 & 0.220 & 0.469 & 44  & 102 \\                
SVDD   & 0.331 & 0.574 & 116 & 658                             
       & 0.300 & 0.543 & 15  & 1441                   
       & 0.730 & 0.832 & 146 & 4904 \\                  
S-SVDD           
       & 0.503 & 0.705 & 176 & 1169                            
			 & 0.440 & 0.649 & 22  & 3890                   
		   & 0.815 & 0.853 & 163 & 10085 \\           
S-SVDDr1            
       & 0.540 & 0.730 & 189 & 1404                            
			 & 0.400 & 0.622 & 20  & 3138             
		 	 & 0.780 & 0.854 & 156 & 6221 \\
S-SVDDr2            
       & 1.000 & 0.000 & 350 & 92685                             
			 & 0.220 & 0.465 & 11  & 1762 
		   & 0.995 & 0.003 & 199 & 92683 \\                
\hline
\end{tabular}
\end{table*}
\vspace{-5pt}
\subsection{Classifiers and their parameters}

As our base-model, we fine-tuned a VGG16 network \cite{Simonyan15} pretrained on ImageNet using FIN-Benthic2 dataset. To make VGG16 suitable for our task, we added two dense layers on top of the VGG16 convolutional output. The first added layer is composed of 4069 neurons with ReLU activation. The second added layer is the output layer composed of 39 neurons using soft-max activation. We also added two dropout layers on top of the mentioned dense layers to avoid overfitting. The dropout rate was set to 40 percent. We fine-tuned the whole network for 50 epochs using Stochastic Gradient Descent with a learning rate of 0.007 and selected the final network based on the validation set accuracy. As the original images are of varying size, we first scaled them to 64x64. The overall accuracy of the network on the test set was 0.872. This is similar to earlier published results \cite{arje2017human}, while we did not concentrate on optimizing this step in this work.

We extracted the output of the second last VGG16 layer (i.e., 4096-d) for further analysis and first applied PCA on it. We used only the target class training samples to obtain the PCA mapping and then applied this mapping for all the remaining data. We kept the first 100 principal components as our final feature vectors used for training and testing the one-class classifiers. Finally, we trained different one-class classifiers (separate models for each target species) using feature vectors of the training images of the target species. The hyper-parameters were optimized using the validation set. At the end, we tested the models with the full test set, where all the images not belonging to the target class were considered as outliers. The one-class classifiers considered were OC-SVM, SVDD, S-SVDD, S-SVDDr1, and S-SVDDr2 (See Section~\ref{ssec:occ}). We used both linear and non-linear (kernel) versions. For the kernel version, we used the RBF kernel, i.e. $\textbf{K}_{ij} = \exp  \left( \frac{ - ||{\textbf{x}_i} -{\textbf{x}_j}|{|^2}}{{2\sigma ^2}} \right)$, where $\sigma$ is an additional hyper-parameter. The hyper-parameters $C$, $d$, $\eta$, $\beta$ and $\sigma$ were selected from the following values: $C\in\{0.01,0.05,0.1,0.2,0.3\}$, $d\in\{1,2,3,4,5,10,20,50, 100\}$, $\eta\in\{10^{-4},10^{-3},10^{-2},10^{-1}\}$, $\beta\in\{0.01,0.1,1,10,100\}$, and $\sigma\in\{10^{-3},10^{-2},10^{-1},10^{0},10^{1},10^{2},10^{3}\}$.

\subsection{Performance metrics}

We report our result in terms of four different criterion: True Positive Rate (TPR) is the fraction of correctly classified target class samples correctly. Geometric Mean (GM) is the square root of the product of TPR and True Negative Rate. GM reflect both the ability of the model to detect target class samples and its ability to keep the overall amount of samples to be manually identified low. Therefore, it was used as our main performance measure used also for optimizing the hyper-parameters. Furthermore, we report the total number of correctly identified target samples i.e., True Positives (TP), and the total number of samples needing manual identification, i.e., True Positives and False Positives (TP+FP). 

\subsection{Experimental results}

We give the experimental results in Table \ref{resultsexperiments}. We see that one-class classifiers, using the same features as VGG16, can indeed detect samples from rare species much better than the deep network with a reasonable overhead (TP+FP). Here, it should be remembered that up to 50 images can represent the same specimen and, therefore, the actual number of specimens needing manual inspection may be significantly smaller than the reported number of images. The best one-class classifier in terms of GM is the linear S-SVDDr2 model. 

\section{Conclusion and future work}
\label{sec:conclusion}

We proposed a taxa identification framework, where specimens potentially representing rare species are directed for human expert inspection. We showed that one-class classifiers can complement a deep neural network with high overall classification accuracy in a way that allows dividing the tasks between machine and human expert. This supports moving from fully manual to semi-automated taxa identification in biomonitoring. The best one-class classification model in terms of Geometric Mean was regularized linear Subspace Support Vector Data Description.

In this paper, we considered images separately, while we actually have multiple images of a single specimen. In our future work, we will consider how to exploit this information. For example, we may require a certain fraction of images to be classified as target class to assign the specimen for human inspection or we may use multi-modal one-class classifiers, e.g., \cite{sohrab2019multimodal}, by considering each image as a separate modality. We will experiment on how to use classification confidences of both the CNN and one-class classifiers to further reduce the number of samples requiring human inspection. We will also experiment with different classifier types, such as class-specific classifiers, in our general identification framework.    
\newpage
\bibliographystyle{IEEEbib}
\bibliography{refs}
\end{document}